\documentclass[11pt]{article}
\usepackage{color}
\usepackage{xcolor}
\definecolor{darkblue}{rgb}{0, 0, 0.5}
\usepackage[colorlinks=true, citecolor=darkblue, linkcolor=darkblue, urlcolor=darkblue]{hyperref}
\usepackage{nodalida2025}
\usepackage{microtype}
\usepackage{times}
\usepackage{url}
\usepackage{latexsym}
\usepackage{graphicx}
\usepackage{booktabs}
\usepackage{amsmath}
\usepackage{amssymb}
\usepackage[nameinlink,capitalize,noabbrev]{cleveref}
\usepackage{nicefrac}
\usepackage{inconsolata}
\usepackage[skip=15pt]{caption}
\usepackage{enumitem}
\usepackage{comment}
\setlist[itemize]{left=0.0cm}
\setlist[enumerate]{left=0.0cm}
\usepackage{adjustbox}
\usepackage{float}
\usepackage{linguex}
\usepackage{multirow}
\usepackage[disable]{todonotes}
\usepackage{nicefrac}

\aclfinalcopy % Uncomment this line for the final submission

\title{Multi-label Scandinavian Language Identification (SLIDE)}

\author{Mariia Fedorova$^*$, Jonas Sebulon Frydenberg$^*$, Victoria Handford$^*$, \\[-0.4ex] \textbf{Victoria Ovedie Chruickshank Langø$^*$, Solveig Helene Willoch, Marthe Løken Midtgaard,} \\[-0.4ex] \textbf{Yves Scherrer, Petter Mæhlum, David Samuel} \\
Department of Informatics, University of Oslo \\
{\tt \{mariiaf,jonassf,vlhandfo,victocla,solvehw,marthemi,} \\[-0.4ex] {\tt yvessc,pettemae,davisamu\}@ifi.uio.no} \\
}

\date{}

\begin{document}
\maketitle
\def\thefootnote{*}\footnotetext{Equal contribution.}\def\thefootnote{\arabic{footnote}}
\begin{abstract}
Identifying closely related languages at sentence level is difficult, in particular because it is often impossible to assign a sentence to a single language. In this paper, we focus on multi-label sentence-level Scandinavian language identification (LID) for Danish, Norwegian Bokmål, Norwegian Nynorsk, and Swedish.\footnote{While acknowledging that the term \textit{Scandinavian} in English sometimes also includes Icelandic and Faroese, we use the term Scandinavian in the sense of \textit{Mainland Scandinavian}, in accordance with established and legal usage of the term in these languages. We also consider Swedish as a single language, overlooking the nuances between Finland-Swedish and Sweden-Swedish.} We present the Scandinavian Language Identification and Evaluation, SLIDE, a manually curated multi-label evaluation dataset and a suite of LID models with varying speed--accuracy trade-offs. 

We demonstrate that the ability to identify multiple languages simultaneously is necessary for any accurate LID method, and present a novel approach to training such multi-label LID models.
\vspace{1.5em}
\end{abstract}

\section{Introduction}

Correctly identifying the language of a short piece of text might seem like a simple (and possibly already solved) task. While differentiating between two distant languages might be straightforward, we show that, when focusing on a group of closely related languages, this task becomes substantially more challenging. This is especially true when we consider the fact that language identification (LID) tools have to be fast and efficient, as they are often used for preprocessing large quantities of texts.

In this paper, we focus on the four closely related Scandinavian languages: Danish, Norwegian Bokmål, Norwegian Nynorsk, and Swedish. In order to accurately differentiate within this group, we move away from the standard single-label (multi-class) language identification and instead treat this problem as multi-label 
classification task, allowing for the identification of multiple languages simultaneously as illustrated in \Cref{fig:multilabel}. Sentences valid in multiple Scandinavian languages are fairly common---they account for about 5\% of our evaluation dataset and 16\% of the sentences shorter than 6 words. If not accounted for, these examples can skew evaluation of existing systems. The three main contributions of SLIDE (Scandinavian Language Identification and Evaluation), are as follows:

\begin{figure}[!t]
    \centering
    \includegraphics[trim={0 0.7em 0 0},clip,width=\columnwidth]{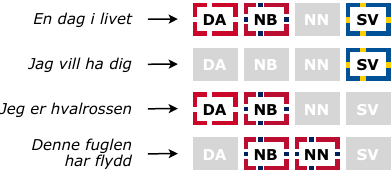}
    \caption{\textbf{Scandinavian similarity}\hspace{1em}Accurate language identification has to necessarily be multi-label when discriminating between closely related languages.}
    \label{fig:multilabel}
\end{figure}

\begin{enumerate}
\item \textbf{A multi-label evaluation dataset}\hspace{1em}We have created a manually corrected multi-label LID dataset for four Scandinavian languages. We present two evaluation methods using this dataset: one designed for a more accurate evaluation of traditional multi-class LID methods, and a second for assessing the performance of multi-label methods.
\item \textbf{A suite of LID models}\hspace{1em}We train a family of language identification models of varying complexities. The best performing models are based on fine-tuned BERT models and smaller, substantially faster models based on FastText embeddings. The source code, datasets and models are released at {\url{https://github.com/ltgoslo/slide}}.
\item \textbf{A novel multi-label LID method}\hspace{1em}Manual creation of a clean multi-label LID dataset is costly. Instead, we present a novel method of silver-labeling such a dataset by utilizing existing machine translation models.
\end{enumerate}

\section{Related work}

\paragraph{Language identification} The task of identifying the language of a text is an ``old'' NLP task dating back to the 1960s. Simple but relatively powerful tools have been available since the 1990s \citep{jauhiainen-jair2019}.

In recent years, the main focus of NLP research has shifted towards large language models, and especially towards extending their coverage to an increasing number of languages. As training data for underrepresented languages is mostly found in web crawls, reliable LID systems covering a large number of languages are more important than ever. While the earliest LID systems were restricted to a dozen languages, recent systems cover hundreds \citep{joulin-etal-2017-bag,grave-etal-2018-learning,burchell-etal-2023-open,heliots2022} and even thousands \citep{kargaran2023glotlid} of languages.

In terms of methods, simple linear classifiers with character-level and word-level features have often outperformed more sophisticated neural models \citep{jauhiainen-jair2019}. Most currently available large-coverage LID models are based on the FastText architecture \citep{joulin-etal-2017-bag}, a multinomial logistic regression classifier with character n-gram embeddings as input features. These include FastText-176 \citep{joulin-etal-2017-bag,grave-etal-2018-learning}, NLLB-218 \citep{nllbteam2022languageleftbehindscaling}, OpenLID \citep{burchell-etal-2023-open} and GlotLID \citep{kargaran2023glotlid}. Different approaches are used by HeLI-OTS \citep{jauhiainen-etal-2022-heli}, which bases its decisions on a combination of character n-gram and word unigram language models, and \texttt{gpt2-lang-ident}\footnote{\url{https://huggingface.co/nie3e/gpt2-lang-ident}}, which is a fine-tuned decoder-only model \citep{radford2019language}.

In practice, LID is most often applied to individual sentences, even though the tools can work with longer or shorter segments of text.

\paragraph{LID for closely related and Nordic languages}

To our knowledge, the only publication focusing specifically on LID for Nordic languages is \newcite{haas-derczynski-2021-discriminating}. They compile a dataset for the six languages (including both Norwegian standards) from Wikipedia and evaluate a range of LID models on it. They find that the languages mostly cluster into three groups: Danish--Bokmål--Nynorsk, Swedish, and Icelandic--Faroese. Their models were not available online as of writing this paper. Besides this, \newcite{nbnordiclid} present two FastText-based LID models: one containing only the 12 most common languages of the Nordic countries (including several Sámi languages, Finnish, and English), and one with an extended coverage of 159 languages.

Futhermore, the previously mentioned off-the-shelf LID systems (NLLB-218, OpenLID, GlotLID, HeLI-OTS) cover all six Nordic languages, with the exception of FastText-176, which does not include Faroese.

\paragraph{Multi-label language identification}

Most existing LID training and evaluation corpora are not manually labeled. Instead, they are based on the assumption that the language is determined by the source it is retrieved from. If a sentence is retrieved from a Danish newspaper, it is assumed to be only Danish. 
But when dealing with closely related languages, it is often the case that an instance cannot be unambiguously assigned to a single language \citep{goutte-etal-2016-discriminating,keleg-magdy-2023-arabic}.

Recent proposals address this issue by framing LID between similar languages as a multi-label task \citep[e.g.,][]{chifu-etal-2024-vardial,abdul-mageed-etal-2024-nadi} and by manually annotating the evaluation data \citep[e.g.,][]{zampieri-etal-2024-language,miletic-miletic-2024-gold}. However, these works do not include studies of Scandinavian languages.
\section{Data}

One of the main contributions of this paper is the release of manually and automatically annotated multi-label datasets. In \Cref{sec:data-sources}, we introduce the sources from which we compile our datasets. 
We then present our manually annotated multi-label evaluation dataset (\Cref{sec:evaluation-data}).
Next, we describe a way to obtain multi-label annotations automatically for the larger training set in \Cref{par:extension}. Lastly, we outline different approaches to data augmentation in \Cref{sec:data-augmentation}.

\subsection{Data sources} \label{sec:data-sources}

As a starting point, we use the Universal Dependencies 2.14 treebanks \citep[UD;][]{nivre-etal-2016-universal, nivre-etal-2020-universal}, keeping their train/dev/test splits intact.\footnote{Specifically, we use the following UD treebanks: 
\href{https://universaldependencies.org/treebanks/no_bokmaal/}{\texttt{no\_bokmaal}}, 
\href{https://universaldependencies.org/treebanks/da_ddt/}{\texttt{da\_ddt}}, 
\href{https://universaldependencies.org/treebanks/no_nynorsk/}{\texttt{no\_nynorsk}}, and 
\href{https://universaldependencies.org/treebanks/sv_talbanken/}{\texttt{sv\_talbanken}}.} 
For each of the four languages, we associate each sentence in the treebank with the language tag corresponding to that treebank's language. This results in a foundational single-label dataset with the following language tags: Danish (\textsc{da}), Norwegian Bokmål (\textsc{nb}), Norwegian Nynorsk (\textsc{nn}), and Swedish (\textsc{sv}). We further incorporate examples labeled as \texttt{other}, drawing random samples from other UD treebanks to represent other languages.

As the UD treebanks are manually annotated, we assume that the texts accurately reflect their corresponding languages. Additionally, the treebanks cover multiple genres, improving the robustness of the models to different text varieties. However, while the resulting dataset is clean, it is not disambiguated. For example, a sentence labeled as Nynorsk is almost guaranteed to be in Nynorsk, but it could also be a valid Bokmål sentence.

\subsection{SLIDE dataset: manually multi-labeled evaluation data} \label{sec:evaluation-data}

\paragraph{Manual inspection} To identify multi-label instances in the validation and test splits, we performed a combination of automatic filtering and manual annotation. Automatic filtration was done by removing frequent words that unambiguously define a language (e.g. `ikkje' is only valid in Nynorsk; the full list is to be found in our Github repository).  

After filtering, we split the remaining instances among a group of annotators to manually check for cases of multilingual acceptability. All annotators were native or near-native Norwegian speakers. Annotation tasks were delegated depending on the speakers' knowledge and exposure to Swedish and Danish (all native speakers have received education in or about other Scandinavian languages through the public curriculum or university classes).

\paragraph{Unclear instances} Most cases of multilingual acceptability involved short sentences with proper names, numbers, or words that are acceptable in multiple Scandinavian languages. Instances consisting of only proper names were annotated with all Scandinavian languages, even if more common in one language than another. Numerical values were treated similarly as they are universally acceptable across the languages.

\begin{table}[]
    \begin{adjustbox}{width=\columnwidth}
    \begin{tabular}{@{}l@{\hspace{4em}}r@{\hspace{2em}}r@{\hspace{2em}}r@{}}
        \toprule
        \textbf{Language} & \textbf{\begin{tabular}[r]{@{}r@{}}Train\\[-0.5em] split\end{tabular}} & \textbf{\begin{tabular}[r]{@{}r@{}}Validation\\[-0.5em] split\end{tabular}} & \textbf{\begin{tabular}[r]{@{}r@{}}Test\\[-0.5em] split\end{tabular}} \\
        \midrule
        Bokmål & 23\,120 & 2\,543 & 2\,098 \\
        Danish & 5\,977 & 563 & 677 \\
        Nynorsk & 21\,587 & 2\,031 & 1\,628 \\
        Swedish & 6\,911 & 553 & 1\,250 \\
        Other & 8\,360 & 1\,124 & 1\,745 \\ \midrule
        
        Total & 61\,406 & 6\,433 & 6 \,950\\ \bottomrule
    \end{tabular}
    \end{adjustbox}
    \caption{\textbf{Dataset sizes}\hspace{1em}Number of sentences per language. Multi-label samples are reported once for each language, while the summary row shows total number of unique sentences.}
    \label{data_distribution}
\end{table}

\paragraph{Non-Scandinavian instances}
Sentences from other languages that are not valid in the Scandinavian languages retain the \texttt{other} label, and we set restrictions on when this label is used. This distinction is crucial as it ensures that the \texttt{other} label exclusively identifies non-Scandinavian sentences, setting it apart from the potential multi-label nature of  the remaining labels. For example, this instance from the Danish treebank, ``- Gerne.", is labeled as only Danish, despite it also being acceptable in German. This approach allows us to evaluate a model's ability to handle ambiguity and focus on the sentences that could belong to multiple Scandinavian languages, without having to consider all possible languages.

\paragraph{Punctuation errors} We found several sentences that were orthographically identical in Danish and Bokmål, where commas were the sole distinguishing factor. When a subordinate clause occurs in the first position of a sentence, both languages include a comma at the end of the clause. However, if the subordinate clause does not occur in the first position, Danish can include a comma before that clause\footnote{\url{https://ro.dsn.dk/?type=rulesearch&side=49}}, whereas Norwegian cannot\footnote{\url{https://sprakradet.no/godt-og-korrekt-sprak/rettskriving-og-grammatikk/kommaregler/}}. The optional comma, in this case, means that Danish can follow the same punctuation rules as Norwegian but does not have to, making differentiation difficult.

Such a sentence is shown in example \ref{comma-ex} from the Danish treebank. The words in this sentence are written the same in Danish and Bokmål however, the comma introducing the subordinate clause \textit{at hun skulle havne på et teater} is technically not allowed in Norwegian. 

\ex.\label{comma-ex} \small{Der stod ingen steder i Mai Buchs eksamenspapirer, at hun skulle havne på et teater.} \\[0.5em]
\small{\textit{It said nowhere in Mai Buch's exam papers that she would end up in a theater.}}

We decided to annotate such sentences as both Danish and Bokmål, thereby focusing on lexical information rather than punctuation. This is  due to Norwegians' challenges with following comma rules in general \citep[pp. 37-39]{michalsen2015komma}, perhaps due to Norwegian earlier having Danish comma rules \citep{papazian2013moltke}. We also find 29444 examples of a comma preceding \textit{at} `that' in the Norwegian LBK corpus, keeping in mind that some of these might be examples of other usage \citep{LBKleksikografiskbk}. 

\paragraph{Code switching} There were also sentences in the dataset that included more than one language. One such example is: 

\ex. {\small Låten heter ``The spirit carries on.''} \\[0.5em]
\small\textit{The song is called ``The spirit carries on.''}

For these sentences that include non-Scandinavian words, we annotated them for the Scandinavian languages only. In cases where a sentence had words from different Scandinavian languages, e.g. a Nynorsk quote in a Bokmål sentence, we made small changes to make the sentence monolingual.\footnote{There were few instances of this, however, it is important to mention that there is not a complete 1-to-1 correlation between the source material and our dataset.}

\paragraph{Number of multi-label instances}
The statistics of the validation and test sets are shown in \Cref{data_distribution}. The resulting shares of multi-label instances in the validation and test sets are 6\% and 5\% respectively. 

\subsection{Automatically multi-labeled training data} \label{par:extension}
As there is no available multi-labeled training dataset for any subset of the Scandinavian languages, and manually annotating a large-enough dataset would be out-of-scope for this project, we decided to silver-label the UD training split automatically. To do so, we converted the task of machine translation into the task of language identification. This conversion then allows us to utilize existing high-quality resources for multi-label language identification.

\paragraph{Machine translation conversion}
The method relies on our observation that machine translation models tend to stay conservative and minimize the changes between the source and target texts. Thus, if the translation of a sentence does not lead to any changes, we label it as a valid sentence of the target language. This means that the machine translation model can only add additional language labels to a sentence as a result; we do not use the translated sentences in any other way.

Specifically, we use \texttt{NorMistral-11b} to perform the translation \citep{samuel2024smalllanguagesbigmodels}. While this large language model is able to translate in a zero-shot manner, we increase its reliability by fine-tuning it on the small high-quality Tatoeba evaluation set \citep{tiedemann-2020-tatoeba} in all translation directions between Bokmål, Danish, Nynorsk and Swedish.

\subsection{Data augmentation} \label{sec:data-augmentation}

\paragraph{Punctuation augmentation} 

\begin{table}[t!]
  \centering 
  \begin{adjustbox}{width=\columnwidth}
  \begin{tabular}{@{}l@{\hspace{3em}}cc@{}}
    \toprule
    \textbf{Alterations} &
    \textbf{\begin{tabular}[c]{@{}c@{}}Loose\\[-0.5em] accuracy\end{tabular}} &
    \textbf{\begin{tabular}[c]{@{}c@{}}Exact-match\\[-0.5em] accuracy\end{tabular}} \\ \midrule
    Augmentation + Regex & 98.6 & \textbf{96.4} \\
    Augmentation & 98.4 & 96.3 \\
    Regex & 98.4 & 96.2 \\
    NER & \textbf{98.7} & 95.5 \\
    Base & 98.3 & 96.2 \\
    \bottomrule
  \end{tabular}
  \end{adjustbox}
  \caption{\textbf{Ablation study}\hspace{1em}Impact of data augmentation and regular expression normalization on SLIDE-base measured by test set performance. "Augmentation" refers to punctuation augmentation, "Regex" refers to regular expression normalization, "NER" refers to named entity swaps and "Base" is neither of the above.}
  \label{tab:ablation}
\end{table}

To prevent our models from relying too much on punctuation, we augment the training data with random punctuation. This is especially important for disassociating punctuation from the \texttt{other} tag, for which the training data exhibits punctuation noise to a higher degree than the Scandinavian language examples. We randomly add either (i) a period, an exclamation point, or a question mark to the end of the sentence or (ii) a hyphen, dash  
or comma at the beginning of the sentence. Additionally, there is a $\nicefrac{1}{3}$ chance of including an intervening space. This augmentation scheme is chosen to try to mimic punctuation variance that is present in sentence-level (parallel) corpora. 

This method is only applied to instances not labeled as \texttt{other} and is performed on about 7.5\% of the training data. This value is heuristically chosen.

\paragraph{Regular expression normalization} 
We normalize URLs, email addresses, and numbers into the following special symbols: \texttt{\textlangle URL\textrangle}, \texttt{\textlangle mail\textrangle{}} and \texttt{\textlangle num\textrangle}. These elements are not informative for language identification, and we do not want a model to associate them with a certain language. 

\begin{figure}[!ht]
    \centering
    \includegraphics[width=\columnwidth]{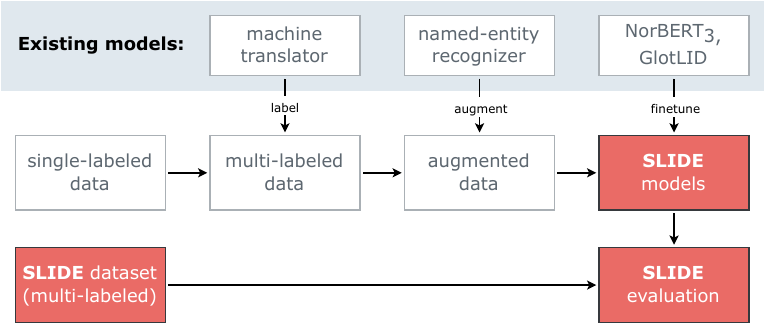}
    \caption{\textbf{Training pipeline}\hspace{1em}A diagram that illustrates the flow of the full training pipeline. We start with a high-quality, single-labeled training dataset, then extend it with multi-label annotations using a strong machine translation model. The dataset is further augmented by randomly swapping named entities identified by existing NER models and through other rule-based augmentations. We use the (augmented) data to fine-tune strong tranformer-based models from a family of NorBERT\textsubscript{3} models \citep{samuel-etal-2023-norbench}, a fast model from the GlotLID static word embeddings. Finally, the manually-annotated multi-label dataset is used to evaluate the resulting models.}
    \label{fig:pipeline}
\end{figure}

\paragraph{Alphabet variations} The alphabet of the four Scandinavian languages differs by the usage of the letters \textit{ä, ö} (in Swedish) and \textit{æ, ø} (in Danish and Norwegian). 
To ensure that the model does not learn to associate the presence of these letters solely with their corresponding languages, we augment the training data by adding Swedish sentences containing the Danish--Norwegian letters and Danish and Norwegian sentences containing the Swedish letters (e.g., in proper names and in the context of quotations).

We use the NPK parallel corpus\footnote{\url{https://www.nb.no/sprakbanken/en/resource-catalogue/oai-nb-no-sbr-80/}} containing translations of news texts from Bokmål to Nynorsk to extract texts containing \textit{ä} and \textit{ö}. For Swedish, we use the EU Bookshop corpus \citep{skadins-etal-2014-billions} to extract Swedish sentences containing \textit{æ} and \textit{ø}. Together, this yielded 10,262 sentences, which are included in \Cref{data_distribution}. 

\paragraph{Named entity swaps} 
We also want to prevent a model from associating named entities with a given language. Although named entities are unequally distributed across languages, they are not necessarily language-dependent. 
We perform named-entity recognition (NER) on the training data using the spaCy\footnote{\url{https://spacy.io/} pipeline. We use the large language-specific models, where the Norwegian model is used for Bokmål and Nynorsk.} to identify and extract persons, organizations, locations, and miscellaneous entities. We randomly swap the recognized entities with other entities from the same category to try to break up any connection between entity name and a given language. 

\section{SLIDE evaluation}

We introduce two evaluation metrics in our comparison: \textit{loose} and \textit{exact-match} accuracy.

\paragraph{Loose accuracy} This evaluation metric is designed for models that output only one language label per input, which is common for off-the-shelf classifiers like FastText and NLLB. According to this metric, a prediction is considered correct if the single predicted label is among the gold labels. This metric is unreliable for multi-label models, since a model that always predicts all four languages would get 100\%.

\paragraph{Exact-match accuracy} This evaluation metric is more strict and requires an exact match between the predicted and gold labels sets, making it more appropriate for models capable of predicting multiple labels. 

\paragraph{Per-language scores}Additionally, we report the F\textsubscript{1}-score for each individual language to measure the quality of classifications for each of the four languages separately. Here, a true positive prediction happens if and only if the respective language is present both in the set of predicted labels and in the set of gold labels.

\section{SLIDE training methodology} 

In this section, we present our approach to training the SLIDE models. We explore two main directions: transformer-based models that achieve high accuracy but require more computational resources, and a fast model based on static word embeddings that trades accuracy for faster inference times. 

\subsection{Transformer models (SLIDE x-small, small and base)} \label{method_transformer}

Fine-tuned masked language models are nowadays the most popular sequence classification solution for problems that require accurate solutions and reasonable inference time \citep{devlin-etal-2019-bert}.

\paragraph{Selection of BERT family}
We assessed massively multilingual, Scandinavian, and Norwegian BERT-like models with comparable number of parameters in order to choose a model to focus on for further optimizations.

We test two massively multilingual models: \textit{XLM-RoBERTa-base} \citep{conneau-etal-2020-unsupervised}, which is trained on a corpus containing 100 languages (including the Scandinavian languages) and has a total of 278M parameters, as well as \textit{DistilBERT-multilingual-base} \citep{Sanh2019DistilBERTAD}, which is a distilled version of the multilingual BERT base model trained on Wikipedia data from 104 languages (including all the Scandinavian languages) with 135M parameters. The Scandinavian model we use is called \textit{ScandiBERT} \citep{snaebjarnarson-etal-2023-transfer}; it is a BERT-like model with 125M parameters trained on Icelandic, Danish, Norwegian, Swedish and Faroese data. 
Finally, \textit{NorBERT\textsubscript{3}-base} \citep{samuel-etal-2023-norbench} is a masked language model trained mostly on Norwegian data.

Preliminary experiments showed that the NorBERT\textsubscript{3} models performed the best on our dataset, as shown in \Cref{tab:size}. We thus use the NorBERT\textsubscript{3} models for further experiments and consider the following sizes from this family of models: \textit{xs} (15M parameters), \textit{small} (40M parameters), and \textit{base} (123M parameters). This allows us to train SLIDE models of varying accuracy-to-speed trade-offs.

\paragraph{Training details}
Fine-tuning is done using the \texttt{transformers} library \citep{wolf-etal-2020-transformers} and the \texttt{PyTorch} framework \citep{Ansel_PyTorch_2_Faster_2024}. We use binary cross-entropy as the loss function to train the model for multi-label classification.

To find our final hyperparameters, we perform a simple grid search. The models are fine-tuned with a learning rate of $5\cdot 10^{-5}$, a batch size of 64, 1\% warmup steps with a linear scheduler together with the AdamW optimizer. We train the models for 3 epochs (2,877 steps) and load the best checkpoint at the end based on metric performance (weighted multi-label accuracy). Model evaluation is performed on the validation set every 100 training steps. We fine-tuned the three NorBERT\textsubscript{3} models in this way and release them as SLIDE-{xs}, SLIDE-small and SLIDE-base.

Various training set compositions were evaluated; the best model was trained on the multi-label UD dataset combined with the `alphabet variations' dataset using the punctuation augmentation approach and regular expression normalization described in \Cref{sec:data-augmentation}. We also observe that lowercasing the training set leads to slightly better performance. Therefore, we applied lowercasing to all the training data. While performance typically improves with more training data, this was not observed on our validation set. The final training set has a skewed label distribution: 35\% Bokmål, 33\% Nynorsk, 13\% other, 11\% Swedish, and 9\% Danish. The validation and test sets reflect similar skews (see \Cref{data_distribution}). 
We briefly tested both upsampling and downsampling to balance labels, but the multi-label nature of the data made this challenging, and it ultimately yielded no improvement. 

\begin{table}[t!]
  \centering 
  \begin{adjustbox}{width=\columnwidth}
  \begin{tabular}{@{}l@{\hspace{3em}}ccc@{}}
    \toprule
    \textbf{Model} &
    \textbf{\begin{tabular}[c]{@{}c@{}}Loose\\[-0.5em] accuracy\end{tabular}} &
    \textbf{\begin{tabular}[c]{@{}c@{}}Exact-match\\[-0.5em] accuracy\end{tabular}} &
    \textbf{\begin{tabular}[c]{@{}c@{}}Macro\\[-0.5em] F\textsubscript{1}\end{tabular}} \\ \midrule
    XLM-RoBERTa-base & 96.8 & 94.6 & 95.4\\
    DistilBERT-base & 96.5 & 94.5 & 95.2\\
    ScandiBERT & 97.6 & 95.9 & 96.6\\
    NorBERT3-base & \textbf{98.6} & \textbf{96.4} & \textbf{97.0}\\
    \bottomrule
  \end{tabular}
  \end{adjustbox}
  \caption{\textbf{Base model selection}\hspace{1em}We made our choice based on the validation data split, the metrics in this table, given in percent, are for the test split. F\textsubscript{1} is per-language exact match. NorBERT3 refers to the same model as SLIDE.}
  \label{tab:size}
\end{table}

\subsection{Static-word-embedding model (SLIDE-fast)} 

Since our dataset is smaller than that used to train baseline FastText models, we train a tiny multi-label model instead of concentrating efforts on pretraining a model on our dataset. The model is based on GlotLID sentence embeddings and has 20.9k parameters, not counting the input embeddings. 
It uses a feed forward network with 1 hidden linear layer of size 64 and a ReLU activation function between it and the output linear layer, and is trained with a regular binary cross-entropy loss. We selected the $0.5$ sigmoid threshold to accept a class based on the validation data split. 
The \texttt{other} class is selected only if all other classes are below the threshold. Reducing number of classes from 2,102 to 4 explains faster inference (\Cref{tab:main-results}) than that of original GlotLID.

\begin{table*}[]
\resizebox{\textwidth}{!}{%
\begin{tabular}{@{}l@{\hspace{4em}}c@{\hspace{0.5em}}c@{\hspace{3em}}ccccc@{\hspace{3em}}c@{}}
\toprule
\textbf{Model} &
  \textbf{\begin{tabular}[c]{@{}c@{}}Loose\\[-0.5em] accuracy\end{tabular}} &
  \textbf{\begin{tabular}[c]{@{}c@{}}Exact-match\\[-0.5em] accuracy\end{tabular}} &
  \textbf{\begin{tabular}[c]{@{}c@{}}\textsc{NB}\\[-0.5em] F\textsubscript{1}\end{tabular}} &
  \textbf{\begin{tabular}[c]{@{}c@{}}\textsc{DA}\\[-0.5em] F\textsubscript{1}\end{tabular}} &
  \textbf{\begin{tabular}[c]{@{}c@{}}\textsc{NN}\\[-0.5em] F\textsubscript{1}\end{tabular}} &
  \textbf{\begin{tabular}[c]{@{}c@{}}\textsc{SV}\\[-0.5em] F\textsubscript{1}\end{tabular}} & \textbf{\begin{tabular}[c]{@{}c@{}}Other\\[-0.5em] F\textsubscript{1}\end{tabular}} & 
   \textbf{\begin{tabular}[c]{@{}c@{}}Runtime\\[-0.5em]ms/sample\end{tabular}} \\ \midrule
\textsc{baselines}  &  &  &  &  &  &  &  \\
\hspace{1em}gpt2-lang-ident & 61.2 & 58.9 & 47.0 & 24.0 & 36.9 & 83.6 & 86.2 & 52.07 \\
\hspace{1em}FastText-176\textsuperscript{*}         & 80.5 & 77.7 &  72.6 & 66.0 & 55.7 & 92.7 & 93.5 & \textbf{0.01} \\
\hspace{1em}NLLB-218\textsuperscript{*}             & 95.3 & 91.6 & 93.0 & 85.9  & 89.0  & 96.8 & 93.6  & 0.08 \\
\hspace{1em}NB-Nordic-LID\textsuperscript{*}        & 83.3 & 80.6 & 85.0 & 67.0 & 84.8 & 89.7 & 70.2 & 0.02 \\
\hspace{1em}OpenLID\textsuperscript{*}              & 94.2 & 90.2 & 91.5 & 82.6 & 88.7 & 95.7 & 93.3 & 0.08\\
\hspace{1em}GlotLID\textsuperscript{*}              & 97.2 & 93.4 &  93.5 & 89.5 & 89.4 & 97.9 & 98.1 & 0.51 \\
\hspace{1em}Heliport (HeLI-OTS)             & 96.5 & 92.6 &  90.9 & 89.0 & 91.2 & 97.6 & 97.2 & 0.02 \\[0.5em]
\midrule
\textsc{our models} &  &  &  &  &  &  &  \\
\hspace{1em}SLIDE-fast   &  95.7  & 93.4 & 94.5 & 90.2 & 92.4 & 97.5 & 96.4 & 0.16\\
\hspace{1em}SLIDE-x-small & 97.8 & 95.7 & 97.5 & 90.4 & 96.2 & 98.0 & 98.7 & 13.22 \\
\hspace{1em}SLIDE-small & 98.1 & 95.7 & 97.7 & 89.9 & 96.3 & 98.0 & 99.1 & 19.70 \\
\hspace{1em}SLIDE-base  &  \textbf{98.6} & \textbf{96.4} & \textbf{98.1} & \textbf{92.0} & \textbf{97.1} & \textbf{98.6} & \textbf{99.4} & 38.41 \\ \bottomrule
\end{tabular}
}
\caption{\textbf{Detailed results on the manually-annotated multi-label SLIDE test split} The best result for each metric is typeset in bold; higher values are always better, except for the runtimes. \textsuperscript{*} shows which baselines use FastText.}
\label{tab:main-results}
\end{table*}

\paragraph{Additional Scandinavian data}

Since a SLIDE-fast model trained on the same training dataset as the larger model does not correctly discriminate Bokmål from Nynorsk and Danish sentences, we enhance the training dataset with additional Bokmål, Nynorsk, Danish and Swedish sentences from the Tatoeba evaluation dataset (automatically labeled in the same way as the UD-based training dataset). NER, punctuation augmentation and regular expression normalization are not applied to the resulting training split.

\section{Experiments}\label{sec:experiments}

We evaluate our SLIDE models against several established LID baselines, comparing both prediction accuracy and speed. Our evaluation focuses on two key aspects: performance on our manually annotated multi-label test set, and generalization to out-of-domain data. We first describe the baseline models used for comparison, then present our main results and the results of our out-of-domain experiments.

\subsection{Baselines}

We compare against LID models available at the time of writing that support the four  Scandinavian languages: FastText-176 \citep{joulin-etal-2017-bag}, NLLB-218 \citep{grave-etal-2018-learning}, NB-Nordic-LID \citep{nbnordiclid}, OpenLID \citep{burchell-etal-2023-open}, GlotLID \citep{kargaran2023glotlid}; Heliport, a faster version of HeLI-OTS \cite{jauhiainen-etal-2022-heli}\footnote{\url{https://github.com/ZJaume/heliport}}, and \texttt{gpt2-lang-ident}.

While top-$k$ prediction with confidence scores is possible for the FastText and GPT2-based models, we observe that the confidence scores are unreliable, i.e. there is no consistent threshold value that improves performance, and for all baseline models, except Heliport, the best results are achieved when they are used as single-label classifiers. 

\subsection{Main results}
\Cref{tab:main-results} presents the main results of our experiments on the manually-annotated SLIDE test set. We report loose accuracy and exact-match accuracy as overall metrics, along with per-language exact-match F\textsubscript{1} scores for each of the four languages and the 'other' category. Additionally, we measure inference speed in milliseconds per sentence, averaged over three runs\footnote{Measured on an AMD EPYC 7702 CPU, with a batch size of 1.}. 

\subsection{Out-of-domain test set} \label{haas_test}

\citet{haas-derczynski-2021-discriminating} provide two test sets with single-label annotations, extracted from Wikipedia. In order to evaluate our models on an out-of-domain dataset and compare them with previous work, we use their two test splits containing 3 000 and 14 960 samples respectively and map Icelandic and Faroese to the `other' label. We present the results on these test sets in \Cref{tab:haas}.

\section{Discussion}

\paragraph{Performance of baseline models} The baseline models exhibit varying levels of performance, see \Cref{tab:main-results} for detailed metrics. These results demonstrate that, while most FastText-based models offer speed advantages, they fall short in accuracy for closely related languages such as Norwegian Bokmål and Norwegian Nynorsk. GlotLID, though slower (0.51 ms/sentence), provides the best performance among the baseline models, with Heliport being a close contender while being significantly faster (0.02 ms/sentence). \texttt{gpt2-lang-ident}, originally pretrained as a monolingual English model,
fails to tell Danish and two Norwegian languages from each other, while being able to detect Swedish and `other', which again highlights the importance of a dataset focused on Scandinavian languages. 

\paragraph{Performance of SLIDE models}
Our three BERT-based LID models SLIDE-xs, SLIDE-small and SLIDE-base perform the best on our test set, with the \texttt{base} version reaching an exact-match accuracy of 96.4\%, while the \texttt{small} and \texttt{xs} both reach 95.7\%. This comes at the cost of significantly longer runtimes compared to the static embedding models. These models are suitable when high accuracy is of most importance. However, it is worth noting that we measured inference speed solely on a CPU, one sentence at a time, to ensure a fair comparison with the faster baseline models intended for CPU usage. Using a GPU with larger batch sizes would result in significantly faster runtimes for the transformer models.

While our SLIDE-fast model reaches the same exact-match accuracy as GlotLID, 93.4\%, it performs better on Nynorsk, Bokmål and Danish, with Nynorsk performance increasing by 3\%. 

Overall, performance on Danish is consistently the lowest---the best model reaches 92\% F\textsubscript{1}. Our models have been trained on more Bokmål than Danish data, and we observe a slight tendency to predict only Bokmål instead of both Bokmål and Danish for multilingual samples. We do, however, notice the same trend with lower Danish performance across all evaluated models, see \Cref{tab:main-results}.

As seen in \Cref{tab:ablation}, the punctuation augmentation led to minor performance improvements. The main motivation behind this approach, however, is increased robustness to noisy data. While the model trained with named entity swapping (see \cref{sec:data-augmentation}) gained the highest loose accuracy performance, 98.7\%, it performed poorly on exact-match accuracy, 95.5\%. We therefore decided not to include this in the final SLIDE models.

\paragraph{Error analysis} Common error sources are proper names (half of `other' instances misclassified as Scandinavian contains proper names (e.g. `kruvi: \textit{Karl Marx}'), instances in English (30\% of `other' instances misclassified as Scandinavian), and loanwords (`- Ta avisa \textit{Kommersant}.', 
`Server med pastasalat med bakte grønsaker og \textit{tsatsiki} til', `Men Anne Linnet - \textit{oh la la}.')
Bokmål and Nynorsk are confused most often. If a sentence valid both in Bokmål and Nynorsk contains irregular Bokmål spelling like `høg' instead of `høy',  and `tjuvfiske' instead of `tyvfiske',
it is likely to be misclassified as Nynorsk only. Some errors imply that particular tokens influence the prediction more than a sentence representation as a whole: `høyre' is a valid word both in Nynorsk (`hear') and Bokmål (`right'), but the sentence `I alle år har vi fått \textit{høyre} at med dagens forbruk er det olje nok for mange tiår.', which is Nynorsk because of `høyre' used as a verb, is misclassified as Bokmål, while a both Bokmål and Nynorsk sentence `I den nye designen er \textit{høgre} og venstre spalte på framsida til nettavisa fjerna.' is misclassified as only Nynorsk because of the spelling. Additionally, some `other' instances containing subwords matching those in Scandinavian are misclassified, although the whole sentence semantics does not make any sense: `Va shiaulteyr \textit{er} \textit{ny} skeabey harrish boayrd.' (Manx).

\paragraph{Out-of-domain evaluation}
In order to ensure that we do not overfit to the UD data, we evaluate our models on the out-of-domain test set presented in \Cref{haas_test}, which was the only LID dataset specific for Scandinavian languages available at the time of writing. While SLIDE-base reaches lower performance than GlotLID on this test set, we must add that this dataset is heavily preprocessed: lower-cased and stripped out of numbers, punctuation signs and some accented characters. We also noticed a fair amount of mislabeled sentences in the dataset, with sentences like ``ou di be t aatm ne en wadi'', ``atahualpa yupanqui'' and ``tromssan ruijansuomalainen yhdistys'' being labeled as Swedish, Danish and Nynorsk, respectively. Furthermore, this dataset contains Icelandic and Faroese as the \texttt{other} languages, which are similar to Nynorsk in many cases. In short, we cannot draw confident conclusions from this result, but it hints at the worst-case performance of our models on out-of-distribution inputs.

\begin{table}[t!]
\begin{adjustbox}{width=\columnwidth}
    \begin{tabular}{@{}l@{\hspace{5em}}cc@{}}
    \toprule
    \textbf{Model} & \textbf{3K test split} & \textbf{15K test split} \\
    \midrule
        SLIDE-base & 92.7 & 95.3 \\
        SLIDE-fast & 85.4 & 88.5 \\
        GlotLID & \textbf{93.0} & \textbf{95.7} \\
    \bottomrule
    \end{tabular}
    \end{adjustbox}
    \caption{\textbf{Performance on an out-of-distribution single-labeled datasets}\hspace{1em}Accuracy on the test sets from \newcite{haas-derczynski-2021-discriminating}. As this dataset is single-label, we consider a prediction to be correct, if one of the predicted languages is correct.}
    \label{tab:haas}
\end{table}

\section{Conclusion}

We release a novel multi-label LID dataset for Danish, Norwegian Bokmål, Norwegian Nynorsk and Swedish with manually annotated validation and test splits. Using machine translation for creating a silver multi-label training dataset from a single-label one has proved to be efficient.  

Although fine-tuning models for a specific data source may be helpful to obtain high performance on a selected test set, such models (especially the FastText-based ones) may be not robust towards the test dataset change. Also, excessive training data preprocessing may lead to performance degradation on data from unknown domains compared with training without any preprocessing.

\section*{Limitations}
We limit ourselves to the larger Scandinavian languages, and include neither the other closely related Nordic languages Faroese and Icelandic (also known as Insular Scandinavian), nor the smaller Scandinavian varieties with a limited written tradition, such as Scanian, Elfdalian and Bornholmsk. We also do not look at other sources of variation, e.g., dialectal, diachronic or otherwise different varieties found in literature or social media.

Another limitation is that while all Norwegians generally understand Swedish and Danish well, as these languages are a compulsory part of the public curriculum, and also teaching languages of Norwegian universities, their productive capabilities are much lower, and there might be cases of mislabeling. 

\section*{Acknowledgments}
We would like to thank Helene Bøsei Olsen and Karoline Sætrum for their work on annotating the initial version of the test set. 
Some computations were performed on resources provided by Sigma2 -- the National Infrastructure for High-Performance Computing and Data Storage in Norway.

~

\bibliographystyle{acl_natbib}
\bibliography{nodalida2025, anthology}

\end{document}